\documentclass[11pt,letterpaper]{article}
\usepackage{emnlp2017}
\usepackage{times}
\usepackage{latexsym}
\usepackage[T1]{fontenc}
\usepackage[utf8]{inputenc}
\usepackage{emnlp2017,amsmath,amssymb,bm,graphicx,array,subfig,float,afterpage}

\emnlpfinalcopy

\date{}

\begin{document}

\title{Handling Compounding in Mobile Keyboard Input}
\author{Andreas Kabel, Keith Hall, Tom Ouyang, David Rybach, Daan van Esch, Fran\c{c}oise Beaufays \\ Google Research \\
  {\tt \{aka,kbhall,ouyang,rybach,dvanesch,fsb\}@google.com}}

\maketitle

\begin{abstract}
  This paper proposes a framework to improve the typing experience of
  mobile users in morphologically rich languages.

  Smartphone keyboards typically support features such as
  input decoding, corrections and predictions that all rely on language
  models. For latency reasons, these operations happen on device, so the
  models are of limited size and cannot easily cover all the words
  needed by users for their daily tasks, especially in morphologically
  rich languages. In particular, the compounding nature of Germanic
  languages makes their vocabulary virtually infinite. Similarly,
  heavily inflecting and agglutinative languages (e.g. Slavic, Turkic or
  Finno-Ugric languages) tend to have much larger vocabularies than
  morphologically simpler languages, such as English or Mandarin.

  We propose to model such languages with automatically selected subword units
  annotated with what we call ``binding types'', allowing the decoder to
  know when to bind subword units into words. We show that this method brings
  around 20\% word error rate reduction in a variety of compounding languages.
  This is more than twice the improvement we previously obtained with a more
  basic approach, also described in the paper.
\end{abstract}

With fast-growing use of mobile devices, offering an efficient and pleasant mobile text input
experience has recently become a topic of great interest to
researchers and technology providers. Speech recognition, for example,
has flourished in the last few years, mostly fueled by the need for
convenient mobile input methods~\cite{schalkwyk2010your}. Likewise, handwriting
recognition has gained more traction, especially in languages with
complex scripts such as Chinese and Indic
languages~\cite{keysers2016multi}.

Relatively speaking, keyboard input has received less attention from
the research community, even though it remains a primary input method
as it is generally considered, whether rightfully so or not, to be the most
straightforward way to input text on a mobile device. As we
will see in this paper, the typing experience of users with
morphologically rich languages may be quite suboptimal if no special care
is taken to model linguistic phenomena such as compounding explicitly.

The main function of a soft keyboard is to decode users' touch inputs
into words and sentences, just like a speech recognizer would decode
input waveforms. Advanced keyboards support ``tap typing'', where
users tap the keys corresponding to the characters of a word, and
``gesture typing'', where they swipe their finger across the layout of
the keyboard.  The keyboard decoder typically also offers
``autocorrections'', ``word suggestions'' (or ``word completions''), and ``next
word predictions'', all of which aim at assisting the user in recovering
from typos and ``fat finger'' errors, and ultimately at reducing the
number of keystrokes required to compose a given text.

Keyboard input decoding is generally performed on device, which imposes strict
constraints on model sizes. Typically an embedded language model is
limited to one to ten megabytes on today's mainstream devices.

In this context, correctly handling morphological variants is
complicated. Human languages display a wide variety of morphological
phenomena which linguists have described and categorized, all resulting in fast
growth of the language's vocabulary size. Model vocabularies are normally
determined by making the simplistic, but usually extremely effective, assumption
that text can simply be split into useful units by using whitespace as a
separator. Big word vocabularies are a problem for the paradigm of small
language models, especially when these models are also expected to provide
enough contextual depth to support features such as next-word prediction.

We propose a framework to handle certain classes of
morphological phenomena, such as compounding, agglutination and
contraction. We model distinct subword units (stems and affixes or
constituents of compounds), and we impose programmatic constraints on how these
can combine. The framework rests on a finite-state transducer (FST)
representation of the underlying decoding models, similar to the
familiar speech recognition FST decoding theory.

The rest of this paper contains more background on keyboard input
(Section~\ref{sec:KeyboardInput}), how keyboard improvements can be
measured (Section~\ref{sec:KeyboardTesting}), an introduction to
morphology (Section~\ref{sec:Morphology}), how it affects keyboard
input and a simple solution to mitigate the issue
(Section~\ref{sec:CompoundingAndKeyboardInput}), and a more generic
solution, with application and results for some compounding languages
(Section~\ref{sec:BindingTypes}).

\section{Keyboard Input}
\label{sec:KeyboardInput}

\subsection{Keyboard Decoding}
\label{subsec:KeyboardDecoding}

A tapped input consists of a time series of touch points, $\bf{x}$,
that encodes the coordinates of the user's key presses. For gesture
input, the input trajectory is sampled, e.g. every 100 milliseconds,
to provide a similar time series. The task of the decoder is to find
the word sequence $\bf{w^*}$ that best matches the input sequence
$\bf{x}$.

The decoder relies on a spatial model to provide a probability
distribution over a set of spatial units for a given touch point, and
a language model to enforce word spellings and respect word sequence
probabilities. The spatial model for tap input is typically a Gaussian
distribution centered on each key center. Gesture inputs instead are
often modeled with the so-called ``minimum-jerk model'' that imposes
smoothness maximization constraints on the input
trajectory~\cite{quinn}. Alternatively, a recurrent neural network
model can be used~\cite{oalsharif2015long}.

Given a spatial model and a language model, the keyboard decoder finds the most
likely word sequence given the input, ${\bf w^*} =  \rm{argmax}_{\bf w} P({\bf w} | {\bf x})$, or ${\bf w}^* = \rm{argmax}_{\bf w} P({\bf x} | {\bf w}) P({\bf w})$ with Bayes rule, which is reminiscent
of the fundamental equation of speech recognition. Accordingly, one
can easily adopt a finite-state transducer (FST) representation as in
automatic speech recognition~\cite{mohri2008}. The details of such an
implementation are not critical to this paper, and one example may be found
in ~\cite{FstPaper}.

\subsection{Keyboard Language Models}
\label{subsec:KeyboardLanguageModels}

Similar to the language models that power embedded speech recognition
systems, keyboard language models are typically n-grams of relatively
low order over a limited vocabulary~\cite{mcgraw2016personalized}. Typical
order of magnitudes may be 3-grams with 64K to 128K words, and a
couple of million n-grams.

Because of keyboard features like suggestions, completions and predictions
where the language model is more prominent than the spatial model, the
language model should be carefully crafted. For example, if a user
gestures ``refereed'', they may expect the keyboard to suggest
``referred'' as an alternative, but not misspellings such as ``referedd'' or
``reffered'' which may appear with non-negligible frequency in the
training corpus, as most corpora contain typos. These four words
have the same spatial score, since the gestures to produce them are
identical; if their language model scores are close enough, one of the
typos may be presented as a suggestion for the original
input. Keyboard users typically find this annoying, as they often view
the keyboard as a lexical reference as well as an input method.

For this reason, keyboard language models are typically trained to a
fixed vocabulary that has been hand-curated to eliminate misspellings,
erroneous capitalizations, and other undesired artifacts.

\section{Keyboard Testing}
\label{sec:KeyboardTesting}

The rigorous testing of a keyboard decoder and its models is
complex. For one thing, keyboard input methods are interactive; users
can tap, select suggestions, backspace, tap again, and so on. In
contrast, speech inputs are continuous, short of occasional
hesitations and dysfluencies. Moreover, it is not easy to
``transcribe'' a sequence of touch points the way one transcribes a
speech waveform or a handwritten sentence: our brains are not
used to performing this task.

For these reasons, the quality of a keyboard is typically assessed
through simulations. Sentences representative of the expected user
inputs are collected in a test set. For each sentence, $\bf{w}$, an
input sequence $\bf{x}$ is simulated using the keyboard
spatial model as a generative engine. The sequence is then decoded,
and the resulting word string $\bf{w}^*$ is compared to the original
string, $\bf{w}$. A ``word error rate'' (WER) can be computed over all
sentences in the test set.

Additional metrics can be tracked to evaluate other facets of the
system, e.g. ``keystroke saving'' for word completion efficiency, next
word prediction accuracy, but we will focus on the WER, which
illustrates well language model quality improvements.

\section{Morphology}
\label{sec:Morphology}

\subsection{Background}
\label{subsec:Background}

Morphology concerns the analysis of words into their
smallest meaningful units. These units are abstract (and may not even
be contiguous, as in circumfixing and ``nonconcatenative'' languages,
but this is beyond the scope of this work). Here, we are interested
primarily in addressing three phenomena.

The first is inflection, where a single word changes forms due to
syntactic usage; for example, in English, verbs inflect with endings
like ``-ing'' and ``-s'' (``eat,'' ``eating,'' ``eats''). This
phenomenon is more prevalent in languages with complex verb
conjugations (e.g. Romance language verb paradigms), languages with
freer word order, where nominal expressions are inflected for case in
addition to number and gender (e.g. the Slavic languages), and even more so
in so-called agglutinative languages like Turkish.

The second phenomenon is contraction, where words, syllables or groups
of words are shortened by ommitting internal letters. For example,
French makes use of elisions to contract pronouns and prepropositions
with verbs (``j'aime'', ``je l'aime'', ``je n'aime pas'').

The third phenomenon is compounding, where two or more words combine
to form a new lexical item; this phenomenon tends to be restricted in
English (e.g. ``lifeboat'') but is highly productive in many
languages, like Dutch, Danish, and German.

By modeling subword units (e.g. stems/affixes or compound constituents) rather
than full words, we can reduce the size of the language models while
keeping their expressivity constant or even allowing their coverage to grow.

\subsection{Compounding Languages}
\label{subsec:CompoundingLanguages}

While our approach is very general, we focus mostly on compounding in
this paper. In English it is possible for multiple free morphemes to
bind together, as in ``flashlight,'' where both
``flash'' and ``light'' are individually meaningful and valid words.
Two-word compounds are a common word-formation strategy in many
languages; in languages such as Sanskrit and many Germanic languages,
multiple-unit compounding is a productive phenomenon.

German words undergo additional changes when compounding
occurs. Because of this, we cannot simply model compounds as the
junction of two words. For example, the words ``Schwan'' (``swan'') and
``gesang'' (``song'') combine to form ``Schwanengesang,'' where ``-en-'' is an
interfix (the so-called ``glue morpheme''). Likewise the ``-s-'' in
``Ordnungssinn'' (``sense of order''), where the lexical components
are ``Ordnung'' and ``Sinn.''  Alternatively, a final weak vowel may
disappear in compounding, as in ``Schulbus'' (``schoolbus''), where
the first morpheme usually shows up as ``Schule'' (``school''). In all
of these cases, because the first morpheme undergoes alteration in
compounding, the surface form would be out-of-vocabulary from
the model's perspective (i.e., ``Schwan'' shows up but not ``Schwanen,''
``Schule'' but not ``Schul,'' etc.).

\subsection{Modeling Morphology}
\label{subsec:MorphologyModeling}

There have been numerous efforts to improve speech recognition through
morphological and subword language models.  This includes early
attempts to incorporate morphological analysis into semantic language
models \cite{elbeze_derouault_1990} as well as work that explicitly
incorporates morphological components into the language model for
decoding \cite{sak_subword_icassp2010,eldesoky-schluter-ney_2010}.
These models typically are of two varieties; improved word modeling by
allowing for richer analysis of words, and subword modeling where
word-forms are split based on a morphologically-inspired analysis
(e.g., stemming where inflectional affixes are split form the stem of
the word form). To our best knowledge, there is no literature on
morphology handling for keyboard input, where as we will see the issue
materializes slightly differently.

\section{Compounding and Keyboard Input}
\label{sec:CompoundingAndKeyboardInput}

The lack of consistent support for compounding makes many ad-hoc
compounds out-of-vocabulary (OOV) in our lexicon and language model
(LM). Typing an OOV compound in a soft keyboard may result in one of
several undesired behaviors. In the best case, if the user taps the
word carefully, the word will be decoded ``literally'' (meaning
letter-by-letter), but will be red-underlined as it is unknown to the
system. Alternatively, if the components of the compound are
in-vocabulary, the word may be split in its components by the
decoder. Worse, if one of the constituents of the compound is OOV, it
will be autocorrected to another word.

The simple word splitting behavior is frequent enough and simple
enough to prevent that we started with a heuristic for this specific
case.

\subsection{Heuristic to Handle Simple Compounds}
\label{subsec:HeuristictoHandleSimpleCompounds}

Simple compound splitting happens when the insertion of a word
separator is the cheapest spatial change that makes the input
in-vocabulary. It will almost always be unacceptable to the user, as
word splitting changes the meaning of the phrase, or even renders it
ungrammatical. To reduce the number of unwanted word splits, we
implemented a postprocessing step in our decoder logic that would try
to identify word splits introduced by the decoder where the user
likely intended a compound (as opposed to missing a white space).

Experimentation converged on the following, highly heuristic prescription:

\begin{itemize}
\item iterate through decoder candidates for the current word (i.e.,
  user input between user-entered word separators) in order of
  descending spatial model score
\item if a two-word candidate is found, insert an additional decoder
  result for the concatenation of the two words and with a slightly
  improved spatial score, and return
\item if a three-,... word candidate is found, return with the result
  list unchanged.
\end{itemize}

The rationale behind these prescriptions is that the LM score for the
two-word candidate is unrelated to the likelihood of the compound, and
that the likelihood of over-generating or `rambling' will increase if
we consider spatially less likely candidates.

For German, we would only consider candidates in which both words were
capitalized (indicating that both are common nouns). When encountering
a two-word candidate with one or both words lower-cased, we would
still consider spatially lower-scoring two-word candidates, but only
if all higher-scoring candidates were case variants.  The generated
candidate would be the concatenation of the first word and the
lower-cased second word. This confined us to the unambiguous case of
nouns composed of other nouns, while still allowing us to find such
candidates in the frequent case of nouns and verbs only differing in case.

For all other languages, we would only consider candidates for which
both words were lower-case, and would not consider candidates of lower
spatial score than the first two-word candidate.

The score boost for undoing a word split has to be tuned
experimentally to find a good compromise between coverage and
precision.  Results with this method are summarized in
Table~\ref{tab:wers:heuristic}.

\section{Binding Types}
\label{sec:BindingTypes}

The simple approach described above is highly heuristic. In this
section, we propose a more principled approach that leverages the
flexibility of the FST decoder.

\subsection{Subword Unit Modeling with Binding Types}
\label{subsec:SubwordUnitModelingwithBindingTypes}

Compounding languages can generate infinitely many distinct words;
traditional word-based n-gram models, however, are designed to handle
similarly unlimited set of word sequences: An English word-based
n-gram model will be able to deal with {\it summer day} when it knows
about {\it winter day} and {\it summer}, yet a strictly word-based
German model would explictly have to know the analogous formation {\it
  Sommertag}.

\begin{table}[thb]
\begin{center}
\begin{tabular}{|l|c|c|c|}
\hline Language & WER  & WER${}_{rw}$ & $\Delta$WER \\ \hline
Danish & 17.1\% & 16.4\% & -3.9\%   \\ \hline
Dutch  & 17.3\% & 16.4\% & -5.0\%   \\ \hline
Finnish & 18.2\% & 15.9\% & -12.4\% \\ \hline
German & 11.3\% & 11.1\% & -1.6\%   \\ \hline
Norwegian & 17.9\% & 16.9\% & -5.6\% \\ \hline
Swedish & 19.6\% & 18.14\% & -7.5\% \\ \hline
\end{tabular}
\caption{Comparison of word error rates between a baseline decoder and
  one with post-decoding compound rewriting step. Each test set
  contains around 20,000 sentences.}
\label{tab:wers:heuristic}
\vspace{-0.3in}
\end{center}
\end{table}

Hence, a natural approach is to map compounds to n-grams, thus
promoting subword units to words, known compounds to known n-grams,
and, crucially, unknown compounds to backed-off n-grams. At training time, we
automatically determine the inventory of subword units using a decompounder,
described in~\cite{macherey}.

It is important that we still distinguish between sequences forming compounds
and sequences of individual words: First, compound parts may not even be words
in their own right, due to interfixes or elisions.  Second, sequences such as
{\it Sommer Tag} can be grammatical, and will in general differ in frequency
from the compound {\it Sommertag}.

In our approach, a subword unit is uniquely specified by its text and
by its 'binding type', comprising two non-negative integers---its left
and right 'binding class', here notated with left subscript and right
superscript.  We require that right and left binding classes of two
consecutive subword units agree. A left (right) binding class of 0
marks a left (right) proper word boundary, all other binding classes
denote composition.

Thus, for an English subword lexicon comprising
$$
\left\{
{{}_0{\text{foot}}^1},
{{}_0{\text{base}}^1},
{{}_1{\text{ball}}^0},
{{}_0{\text{foot}}^0},
{{}_0{\text{base}}^0},
{{}_0{\text{ball}}^0}
\right\},
$$ the rules would allow ${}_0{\text{base}}^1_1{\text{ball}^0}$ and
${}_0{\text{foot}}^1_1{\text{ball}}^0$, and forbid all variants of
${}_0{\text{base}}^k_0{\text{base}}^m$ but $k=m=0$ (i.e., the two-word sequence {\it base base}.)

With one inner binding class, a token can be classified as
${{}_0{\text{word}}^0}$, ${{}_0{\text{prefix}}^1}$,
${{}_1{\text{infix}}^1}$, or ${{}_1{\text{suffix}}^0}$. Adding binding
classes $>1$ allows for more fine-grained control, or even for the
addition of different morphological phenomena. For example, adding
$$\left\{{{}_0{\text{un}}^2}, {{}_2{\text{usual}}^0},
{{}_2{\text{happy}}^0}\right\}$$ to the above would capture adjective
negation, while avoiding ${}_0{\text{un}}^2_1{\text{ball}}^0$.

For a given set of such subword units with binding classes $0..N$, a
lexicon FST (i.e., a FST mapping keys to sequences of subword ids) can
be constructed the following way: Let $L_s$ be a subword unit
key-to-word-id FST and $R^{l,r}_{(N)}$ an acceptor for subword units
of binding type $(l,r)$.  By iterating
$$ R^{l,r}_{n-1} = R^{l,n}_{n} + R^{l,n}_n (R^{n,n}_n)^\ast
R^{n,r}_n,$$ (using $+$ for union and an obvious notation for
concatenation and Kleene star) one obtains an acceptor for all
single-word subword sequences: $W = R^{0,0}_0$, and the word-level
lexicon FST is the composition $L_w = (L_s L_s^\ast) \circ W$.

By construction, the subword id sequences generated by this lexicon
FST will be well-formed: traversing the FST once will result in a
complete, rule-conforming sequence spanning exactly one proper
word. Furthermore, it will generate {\it all} such sequences that are
compatible with its input; whether one of these sequences has been
seen in training will be reflected in their LM score. Note that, in
general, $L_w$ will be non-functional, reflecting the fact that
compound analysis need not be unique, as in German {\it Staubecken},
which may be analyzed as {\it Stau} + {\it Becken} (dam reservoir) or
{\it Staub} + {\it Ecken} (dusty corners).

For all languages we investigated, the resulting lexicon was somewhat
smaller (by a few percent) than a lexicon for a purely word-based
model, all other training parameters kept unchanged.

Note that we are not restricted to the simple ruleset above; it was
chosen because it is easy to enforce at training and decoding time and
kept changes to our training pipeline minimal, while being flexible
enough to capture a variety of word formation mechanisms (simple
prefixes/suffixes as well as potentially infinite sequences as in
compounding). In principle, however, we could construct a lexicon
using any regular language over subword classes.

Our scheme is easily implemented in LM training, as compound analysis is a
preprocessing step in our pipeline; the compound parts are annotated in-text
with unique affixes representing their binding type. The subsequent steps of the
pipeline will consider them distinct words, and can remain unchanged. The token
affixes can be stripped when the actual textual value is needed (lexicon
building, decoder user interface).

The resulting LM, however, cannot be used as-is.  Suppose the LM FST
scores the last token in a sequence ending in a compound bigram
${}_0a^1{}_1b^0$, and that bigram has not been seen in training (or
has been pruned from the LM). The LM will successively back off to shorter context
states, eventually from $[{}_0a^1{}]$ to the empty context $[]$. This
state will have outgoing arcs for all pseudo-unigrams, words and
subwords alike, with weights given by their background probabilities.

This vastly underestimates the actual likelihood of ${}_1b^0$ occurring after
${}_0a^1$: at both training and decoding time, its occurrence is conditioned on
being preceded by a token of right binding class 1, while the unmodified LM will
give the unconditional probability.

The remedy is to normalize the probabilities on arcs leaving the empty context
state separately for each left binding class (which is equivalent to introducing
per-binding class back-off states, $[\cdot^0]$, $[\cdot^1]$,\ldots). This
renders the LM probabilistically correct: the probability sum over all {\it
admissible} paths through the LM now is 1.

Realizing this was a crucial step: without this probabilistic correction, our
German model suppressed compound back-off by $\approx e^{3.8}$, effectively
restricting it to decoding/predicting compounds seen at training time.

\subsection{Experiments}
\label{subsec:Experiments}

We ran experiments in Danish, Dutch, and German. We first
trained baseline word-based language models using large corpora of
anonymized messages and web documents. We then trained subword models
from the same data, using in a preprocessing step the compound
splitting library described in~\cite{macherey}, which was modified to
further split compound parts into base word and interfix.  The
clean-up step of our pipeline, which marks out-of-vocabulary words by
comparing against curated vocabularies, was modified to reject a
split-compound n-gram only if the compound itself and all its
constituents (base words, no interfix) were OOV. In that case, the
compound was replaced with a single OOV token. Otherwise, interfixes
were re-attached to their prefix, and the compound parts were
annotated as described above, using a single inner binding class.

All pruning parameters were left unchanged between the baseline and
the experimental configuration: 150k unigrams and 1.5M n-grams, with
maximal n-gram order 3.  The lexicon and LM FSTs were generated as
described above. Table~\ref{tab:wers_ime} gives word error rates with
both approaches.

\begin{table}[htb]
\begin{center}
\begin{tabular}{|l|c|c|c|c|}
\hline Language & WER${}_w$ & WER${}_{sw}$ & $\Delta$WER & $\Delta$Ins \\ \hline
Danish & 17.8\% & 14.2\% & -20.2\% & -69.7\%  \\
Dutch  & 16.7\% & 13.9\% & -16.1\% & -75.0\%  \\
German & 10.6\% & 8.5\% & -19.8\%  & -71.4\%  \\ \hline
\end{tabular}
\caption{Comparison of word error rates between word-based and
  subword-based language models. The relative change in insertion rate
  is a measure of reduction in erroneously split compounds.}
\label{tab:wers_ime}
\vspace{-0.3in}
\end{center}
\end{table}

A more in-depth analysis of decoding results highlighted two
mechanisms contributing to the WER reduction over our previous,
heuristic solution:

(1) compounds with interfixes: German {\it Versicherung} (`insurance')
  becomes the non-word $\ast${\it Versicherungs-} when used as a compound prefix,
  making such compounds inaccessible for our heuristic trick of undoing the
  insertion of a word separator.  It shares this property with all
  other verb nominalizations formed as stem + {\it-ung}. All such
  compounds had to be contained as words in our (purely enumerative)
  lexicon. Thus, we were able to decode {\it Versicherungsvertreter}
  (`insurance salesman'), but not {\it Versicherungsvertreterin}
  (`insurance saleswoman'); {\it Versicherungsbetrug} (`insurance
  fraud'), but not {\it Versicherungsbetrüger} (`insurance
  fraudster').

(2) multi-word compounds: An actual example from our test sets is
  German {\it Modelldampfmaschinenhändler}, (`model steam engine
  dealer').  The word-separation heuristic fails, as there are no
  two-word analyses of this ad-hoc formation: the individual words as
  well as the compound `steam engine' are in the lexicon, but the
  three-word compounds `model steam engine' or `steam engine dealer'
  are not---if they occurred in the corpus at all, they are too
  infrequent. The remaining two-compound analysis {\it Modelldampf} +
  {\it Maschinenhändler} would require a nonsensical lexicon entry for
  `model steam'.

Our subword-based lexicon, however, maps the compound to the four-part sequence $({}_0{\text{\it
      Modell}}^1_1{\text{\it Dampf}}^1_1{\text{\it
      Maschinen}}^1_1{\text{\it Händler}}^0)$.
 When scoring this, the LM backs off to
  the unigram level, except for $({}_1{\text{\it
      Maschinen}}^1_1{\text{\it Händler}}^0)$.  In a nice example of the generalization capabilities of a subword-based LM, the presence of that
  bigram can be traced back to {\it Landmaschinenhändler} (`farm
  machinery dealer') in our corpus, and of ${}_1{\text{\it Dampf}}^1$,
  to {\it Quecksilberdampflampe} (`mercury-vapor lamp').

\section{Conclusion}
\label{sec:conclusion}

We proposed a generic solution to handle a variety of morphological
phenomena as they materialize and affect users in soft keyboard
input. We applied this approach to the case of compounding languages,
and showed that it can drastically reduce the undesired splitting of
compound words, and improve WERs by roughly 20\% across
languages. In future work, we will generalize our decompounding models
to more languages and phenomena and hence extend the reach of the
proposed method.

\bibliography{emnlp2017}
\bibliographystyle{emnlp_natbib}

\end{document}